\pdfoutput=1
\documentclass[11pt]{article}

\usepackage[final]{ACL2023}
\usepackage{times}
\usepackage{latexsym}
\usepackage[T1]{fontenc}
\usepackage[utf8]{inputenc}

\usepackage{microtype}

\usepackage{inconsolata}
\usepackage{graphicx}
\usepackage{floatrow}
\newcommand{\myexp}{\mathsf{e}}

\usepackage{amsmath}
\usepackage[nameinlink,capitalise]{cleveref}
\usepackage{hyperref}
\usepackage{caption}
\title{From \mbox{N-grams} to Pre-trained Multilingual Models For Language Identification}

\author{Thapelo Sindane$^{1}$ \and Vukosi Marivate$^{1,2}$ \\
        $^1$Data Science for Social Impact,\\ University of Pretoria, South Africa \\  
        $^2$Lelapa AI\\
        sindane.thapelo@tuks.co.za,\\ vukosi.marivate@cs.up.ac.za}
\begin{document}  
\maketitle
\begin{abstract}
In this paper, we investigate the use of \mbox{N-gram} models and Large Pre-trained Multilingual models for Language Identification (LID) across 11 South African languages. For N-gram models, this study shows that effective data size selection remains crucial for establishing effective frequency distributions of the target languages, that efficiently model each language, thus, improving language ranking. For pre-trained multilingual models, we conduct extensive experiments covering a diverse set of massively pre-trained multilingual (PLM) models -- mBERT, RemBERT, XLM-r, and Afri-centric multilingual models -- AfriBERTa, Afro-XLMr, AfroLM, and Serengeti. We further compare these models with available large-scale Language Identification tools: Compact Language Detector v3 (CLD V3), AfroLID, GlotLID, and OpenLID to highlight the importance of focused-based LID. From these, we show that Serengeti is a superior model across models: \mbox{N-grams} to Transformers on average. Moreover, we propose a lightweight BERT-based LID model (za\_BERT\_lid) trained with NHCLT + Vukzenzele corpus, which performs on par with our best-performing Afri-centric models.     
\end{abstract}
   
 % \Keywords{Language Identification, \mbox{N-grams}, Naive Bayes, Transformers, Pre-trained Multilingual Models,} 

% \maketitleabstract

\section{Introduction}
% https://github.com/PaulSudarshan/Language-Classification-Using-Naive-Bayes-Algorithm/blob/master/Language_Classifier.ipynb
Automatic language identification (LID) is the task of determining the underlying natural language used in a written or spoken corpus \cite{mcnamee2005language}. This is a challenging problem, especially for languages with insufficient training examples and closely related languages, particularly low-resourced languages \cite{haas2020discriminating}. For South African languages, building quality LID technologies is significantly important for sourcing internet data, which has served as a de-facto repository for many low-resourced languages, especially from public domains such as news websites \cite{marivate2020improving, adelani2021menyo, dione2023masakhapos, adelani2023masakhanews,  lastrucci2023preparing}. 

Statistical approaches for automatic LID such as \mbox{N-grams} \cite{dube2019language}, and more classical machine learning models such as Logistic Regression, Naive Bayes, Random Forest, Boosting machines, Support Vector Machines, and Clustering techniques (e.g K Nearest Neighbors) have been proposed \cite{haas2020discriminating}. Moreover, contemporary neural-based architectures such as deep neural networks and convolutional neural networks have also been tested. In all cases, not enough work for the South African languages is reported. 

On the other hand, recent algorithmic advancements such as transformer architectures have made a significant impact on the Natural Language Processing landscape \cite{devlin2018bert,conneau2019unsupervised}. With this sudden shift in perspective, many works have proposed automatic LID using large pre-trained multilingual models, derived from attention mechanisms \cite{vaswani2017attention}. Large pre-trained multilingual models are transformer-based architectures simultaneously trained on multiple languages (hence multi-lingual) using various techniques such as token (s) masking training technique, where tokens from a given sentence example are hidden and the objective of the training transformer is to predict the hidden word (s). 

In this work, we make use of the recently released Vuk'zenzele crawled corpus \cite{lastrucci2023preparing} and the NCHLT dataset \cite{eiselen2014developing} to develop and experiment on automatic language identification models on 10 low-resourced South African languages: Northern Sesotho (nso), Setswana (tsn), Sesotho (sot), isiZulu (zul), isiXhosa (xho), isiSwati (ssw), isiNdebele (nbl), Tshivenda (ven), Xitsonga (tso), and Afrikaans (af). Additionally, we included the high-resource South African English (eng) to ensure representation of all 11 official languages in South Africa. We conduct extensive experiments on N-gram models, large pre-trained multilingual models -- XLM-r, mBERT, and Afri-centric multilingual models -- AfriBERTa, Afro-XLMr, AfroLM, and Serengeti. We shed light on the limitations and robustness of N-gram-based approaches and the significant improvement boost of pre-trained multilingual models, especially, for those pre-exposed to low-resourced South African languages during pre-training.

\section{Related Work}
Large pre-trained multilingual models have shown astonishing state-of-the-art results on various Natural Language Processing (NLP) tasks such as Machine Translation, Question Answering, and Sentiment Analyses \cite{stickland2020recipes,yang2019end, adebara2022serengeti}. A precursor of these tasks is the crawling of large volumes of internet data and categorizing the data into different languages (i.e. language identification) for pre-training. For language identification, many works have used pre-trained multilingual models to expand monolingual datasets using the internet. 

\citet{jauhiainen2021comparing} conducted a comparative study between adaptive Naive Bayes, HeLI2.0, multilingual BERT, and XLM-r models for Dravidian language identification in a code-switched context (i.e. a conventional modus operandi for communication on the internet). \citet{caswell2020language} developed a transformer-based LID model aside from basic filtering techniques such as tunable-precision-based filtering using a created wordlist, TF-IDF filtering, and a percent-threshold filtering threshold proposed in their study to filter noisy web-crawled content. Although they were able to collect corpora for over 212 languages, their set-up for their best-performing transformer model was unclear. Similar to our work, \citet{kumar2023transformer} conducted a comparative study on DistilBERT, ALBERT, and XLM-r and showed that a lightweight version of DistilBERT delivers comparable results to resource-intense models. \citet{adebara2022afrolid}, on the other hand, implemented a massive transformer-based LID model with 12 attention layers and heads. They then trained this model on 512 languages with close to 2 million sentences across 14 language families (South African languages included). Their model achieved over 95 $\%$ F1 score on a left-out test sets, outperforming available LID tools: CLD version 2, Langid, Fasttext, etc. \citet{kargaran2023glotlid} created a language identifier that covers a whopping 1600 low-high-resourced African languages. Due to the unavailability of resources utilized in previous studies, our research concentrated exclusively on 11 South African languages, with only 3 language families - Sotho-Tswana, Nguni, and Creole. Furthermore, we will only consider a comparison of diverse pre-trained multilingual models (E.g mBERT, XLMr, AfriBERTa, Afro-XLMr, Serengeti, e.t.c) and two lightweight BERT-based models -- DistilBERT, and za$_{-}$BERT$_{-}$lid model. 

% \cite{kreutzer2022quality} argues available automatic LID for African languages to be unreliable.
% \

\section{Methodology}
The methodology employed in this study uses language-identifiable monolingual corpora from reliable sources as training examples for language identification and compares various pre-trained multilingual models for the task of discriminating between languages.
\subsection{Corpora}
\begin{figure}[!ht]
\begin{center}
\includegraphics[scale=0.45]{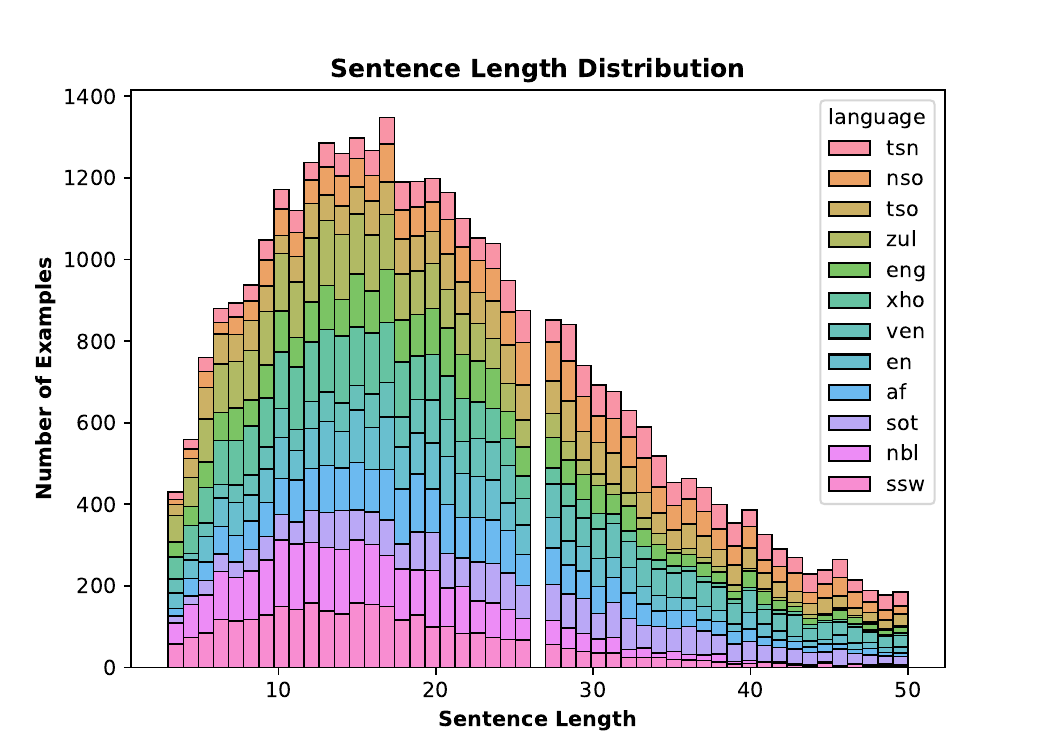} 
\caption{Sentence length distribution of Vuk corpora. The x-axis denotes the number of tokens (words) in the sentences.}
\label{vuksentdist}
\end{center}
\end{figure}

\begin{figure}[!ht]
\begin{center}
\includegraphics[scale=0.45]{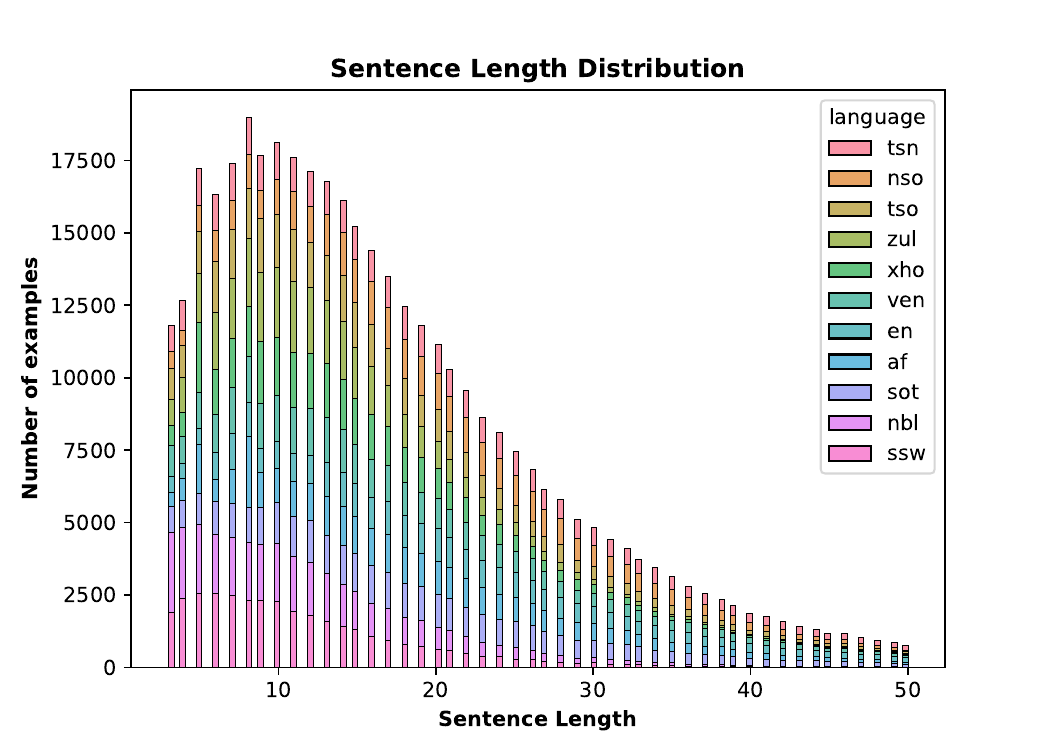} 
\caption{Sentence length distribution of NCHLT + Vuk corpora. The x-axis denotes the number of tokens (words) in the sentences.}
\label{nchltvuksentdist}
\end{center}
\end{figure}

Text corpora for the 11 South African languages were acquired from two sources: Vuk'zenzele (Vuk) \cite{lastrucci2023preparing} and National Centre for Human Language Technology (NCHLT) corpora \cite{eiselen2014developing}. Table \ref{corpora-statistics}, describes the number of sentences (No. Sent), vocabulary (Voc) sizes, unique vocabulary sizes (Unq. Voc), and the train size per language, development set size, and test size per language splits for corpora Vuk and NCHLT. We ensure consistent train and test examples across all languages, by ensuring that all train, and test examples for each language are equal. Therefore, we only had varying development sizes. Additionally, we only considered sentences in the range of 3-50 tokens and did not use the rest of the corpus. Figure \ref{vuksentdist}, and \ref{nchltvuksentdist}, describe the sentence length distribution for Vuk, and NCHLT corpora respectively.

\begin{table}
\tiny
\centering
\begin{tabular}{lllllll}
\hline
\textbf{Corpora} & No. Sent & Voc & Unq. Voc & Train & Dev & Test \\
\hline
Vuk & 33K& 690K & 132K & 3395 & - & 728 \\
% NCHLT & & & & & &  \\
NCHLT + Vuk & 74K & 16M & 258K & 6790 & - & 1454 \\
\hline
\end{tabular}
\caption{\label{corpora-statistics} Corpora statistics for Vuk and NCHLT}
\end{table}

% ------------------- Vuk data -------------

% ------------------- Vuk + NCHLT data -------------

\subsection{Pre-processing}
The dataset is observed to contain links, digits and therefore our pre-processing included the removal of URLs, digits, punctuations, and followed by lower-casing all sentences using Python regular expressions. Special characters such as \v{s}, found in Northern Sotho were left intact.
\subsection{Language detection algorithms}
\subsubsection{\mbox{N-grams}}
\begin{figure}[!ht]
\centering
\includegraphics[scale=0.45]{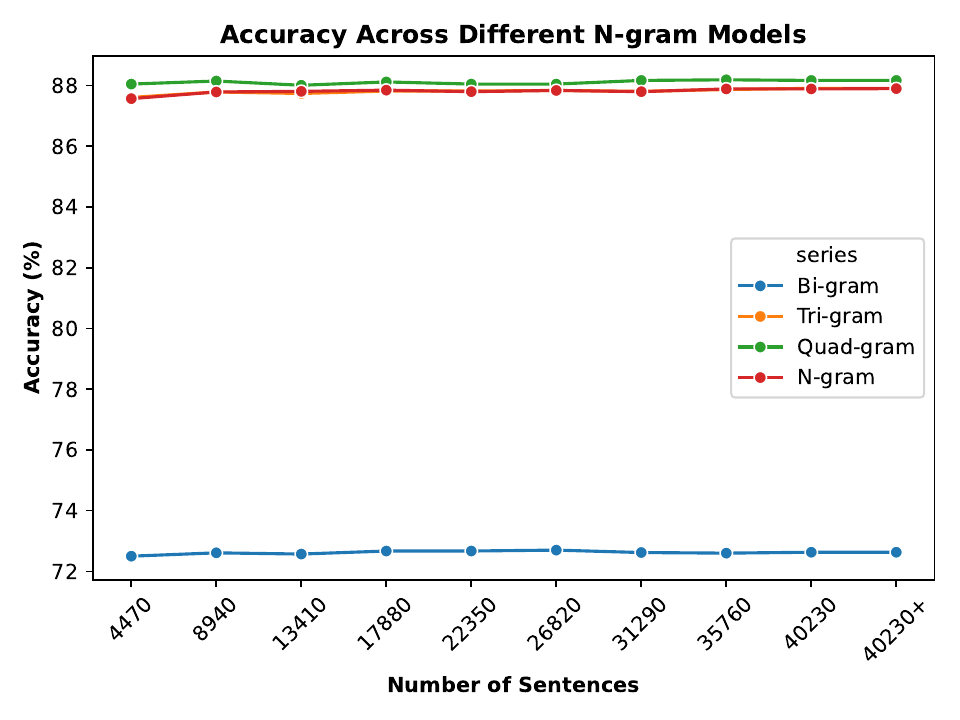}
\caption{Data size variation performance on Vuk test data.}
\label{ablationonvuk} % Place \label here after \caption
\end{figure}

An N-gram is a sequence of consecutive characters from text \cite{dube2019language}. This study explored character Bi-grams (2 consecutive characters), Tri-grams (3 consecutive characters), and Quad-grams (4 consecutive characters) models. We build each model for each language from the training dataset (Vuk, NCHLT, and Vuk $+$ NCHLT). Furthermore, we experimented with various data sizes to investigate the impact of the number of training examples on N-gram models and this showed a performance ceiling, where an increase in training examples does not significantly impact the quality of the models (shown in Figure \ref{ablationonvuk}). Each model is made up of a list of tuples of characters-frequency pair ordered in descending order. 
% --------------------Bi-Figure---------------------

\begin{figure*}[h]
% \hspace*{-1.5cm} % Adjust the value as needed to reduce the left margin
\floatsetup{captionskip=12pt} % Adjust the value as needed
\centering
    \begin{subfloatrow}[3]
        \centering 
        \ffigbox[\hsize]{\caption{Score heatmap for all predictions using N-gram}  \label{figcs1}}{\includegraphics[width=0.3\textwidth]{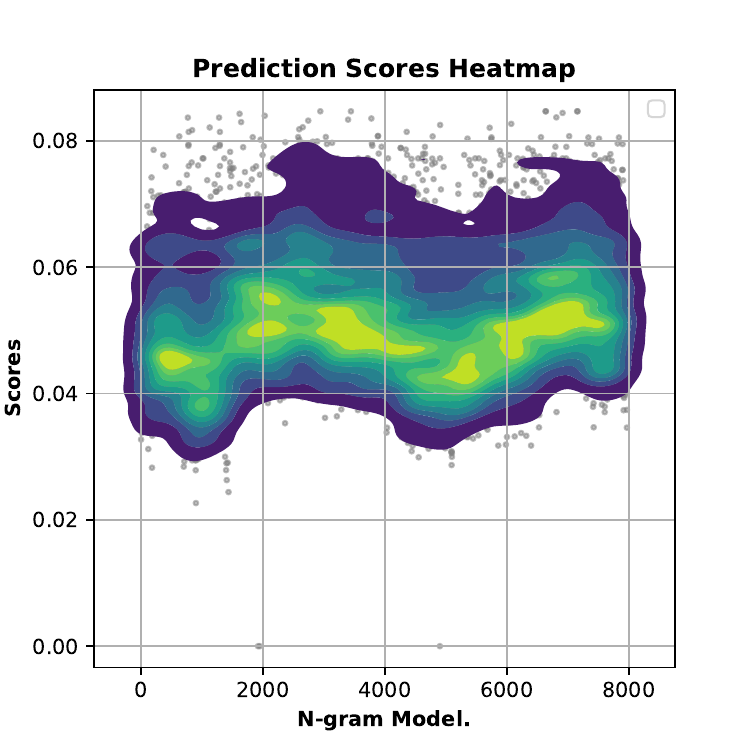}} 
        \ffigbox[\hsize]{\caption{Score heatmap for correctly predicted examples using N-gram} \label{figcs2}} {\includegraphics[width=0.3\textwidth]{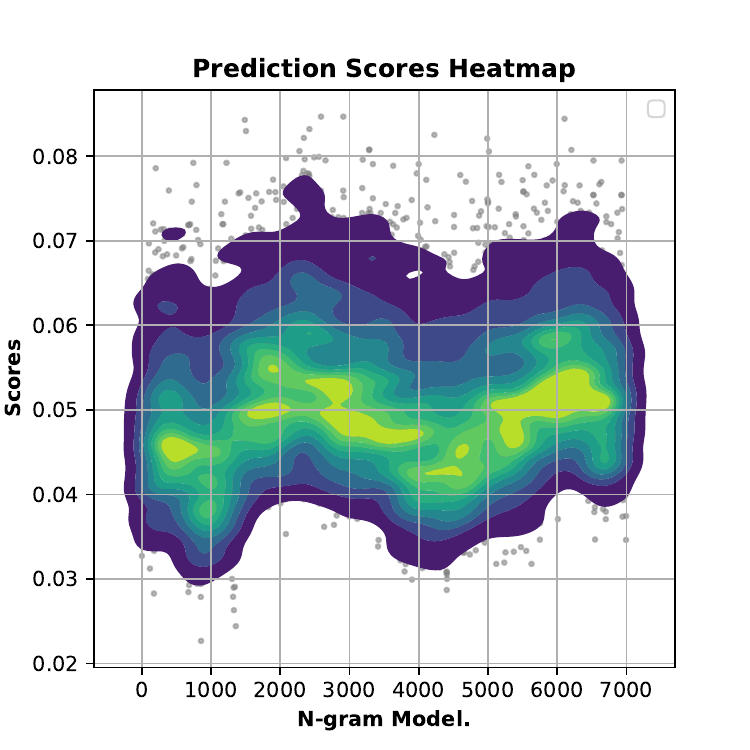}} 
        \ffigbox[\hsize]{\caption{Score heatmap for incorrectly predicted examples using N-gram}\label{figcs3}}{\includegraphics[width=0.3\textwidth]{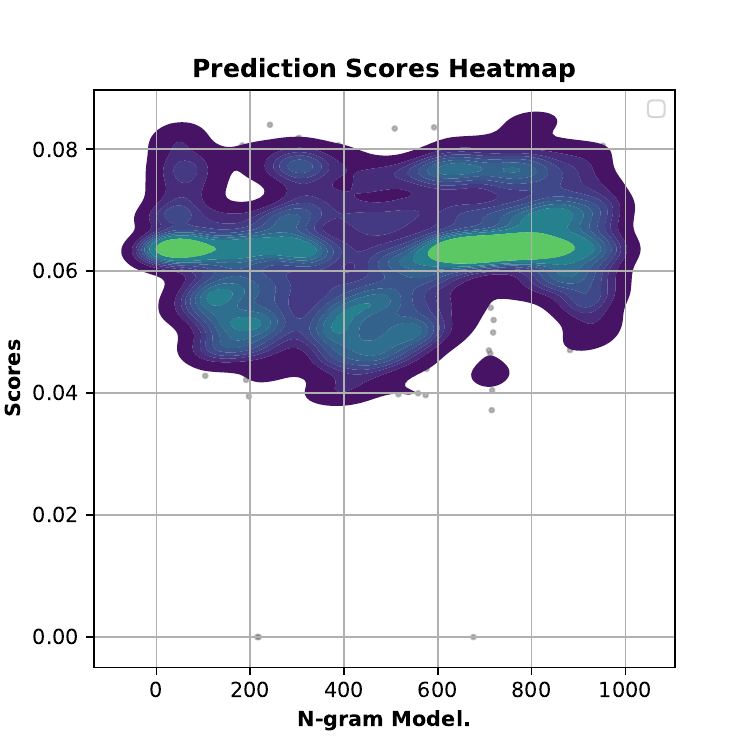}}
    \end{subfloatrow}
    \caption{Comparison of cross-lingual transfer models.}
\label{scoresheatmap}
\end{figure*}

To discriminate between languages, the models use a ranking function. The ranking function calculates the distance of the frequency distributions of the input examples from the existing N-gram model's frequency distributions (with $k$=50 as the number of ordered \mbox{N-grams} to consider from the trained \mbox{N-grams}). The frequency distribution is calculated as the number of occurrences of each observed N-gram divided by the total number of \mbox{N-grams} from the corpus and taking the log of that ratio. For a given input example (in Northern Sotho) "Ke ya go thopa sefoka" translation - "I am going to win the trophy", the model first extracts the character \mbox{N-grams} (e.g. 2 characters if the observed model is Bi-gram) -- Bi-gram Output: ['ke', 'ya', 'a$_{-}$', 'go', 'th', 'ho', 'op', 'pa', 'se', 'ef', 'fo', 'ok', 'e$_{-}$', '$_{-}$y', '$_{-}$g', 'o$_{-}$', '$_{-}$t', '$_{-}$s'] sorted in reverse, and then the frequency distribution from the existing trained models (looking only at 50 top \mbox{N-grams} per language) for all the languages are compared with the new frequency distribution of the input sentence and the one with the closest similarity is considered the language of the input example. Figure \ref{figcs1}, \ref{figcs2}, and \ref{figcs3} presents heatmaps depicting the probability scores generated by the ranking function exclusively for all test examples, correctly predicted sentences, and incorrectly predicted examples, during the test phase respectively. The heatmaps reveal that the concentration of scores ranges between 0.04 and 0.06, which could be further used to drive a model's outcome improving the confidence in predictions. This observation suggests that ranking functions play a crucial role in N-gram-based models, warranting further investigation.

\textbf{\mbox{N-grams} experimental setup}
We experimented with Bi (2), Tri (3), and Quad (4) consecutive character sequences to build our models. Additionally, we combined all 3 and called it \mbox{N-grams} combined.
% ----------------------- HeatMap -------------
% \begin{figure*}[h]
%     \centering
%     \begin{minipage}[b]{0.3\textwidth}
%         \centering
%         \includegraphics[width=\textwidth]{images_and_plots/all_pred_only_score_plots.pdf}
%         \caption{Caption 1}\label{scoresheatmap1}
%     \end{minipage}
%     \hfill
%     \begin{minipage}[b]{0.3\textwidth}
%         \centering
%         \includegraphics[width=\textwidth]{images_and_plots/correct_score_plots.pdf}
%         \caption{Caption 2}\label{scoresheatmap2}
%     \end{minipage}
%     \hfill
%     \begin{minipage}[b]{0.3\textwidth}
%         \centering
%         \includegraphics[width=\textwidth]{images_and_plots/incorrect_score_plots.pdf}
%         \caption{Heatmaps - all predictions (left), correctly predicted (middle), and incorrectly predicted (right)}    \label{scoresheatmap3} 
%     \end{minipage}
%     \caption{Heatmaps showing all predictions (left), correctly predicted (middle), and incorrectly predicted (right).}
%     \label{scoresheatmap} % Label for the entire figure
% \end{figure*}

% \begin{figure}[htbp]
%     \centering
%     \begin{subfigure}[b]{0.35\textwidth}
%         \centering
%         % \includegraphics[width=1.1\textwidth,height=1.8in]
%          \includegraphics[width=\linewidth,height=\linewidth]
%         {images_and_plots/all_pred_only_score_plots.pdf}
%         \caption{CCA cosine scores}
%         \label{figcs1}
%     \end{subfigure}\hspace{-1em}%
%     \begin{subfigure}[b]{0.35\textwidth}
%         \centering
%         % \includegraphics[width=1.1\textwidth,height=1.8in]
%         \includegraphics[width=\linewidth,height=\linewidth]
%         {images_and_plots/correct_score_plots.pdf}
%         \caption{VecMap cosine scores}
%         \label{figcs2}
%     \end{subfigure}\hspace{-1em}%
%      \begin{subfigure}[b]{0.35\textwidth}
%         \centering
%         % \includegraphics[width=1.1\textwidth,height=1.8in]
%         \includegraphics[width=\linewidth,height=\linewidth]{images_and_plots/incorrect_score_plots.pdf}
%         \caption{Muse cosine scores}
%         \label{figcs3}
%     \end{subfigure}
%     \vspace{-12mm}
%     \caption{Comparison of cross-lingual transfer models.}
% \end{figure}
\subsubsection{Naive Bayes Classifier}
\begin{figure}[!ht]
    \centering
    \ffigbox[\FBwidth]{\includegraphics[scale=0.5]{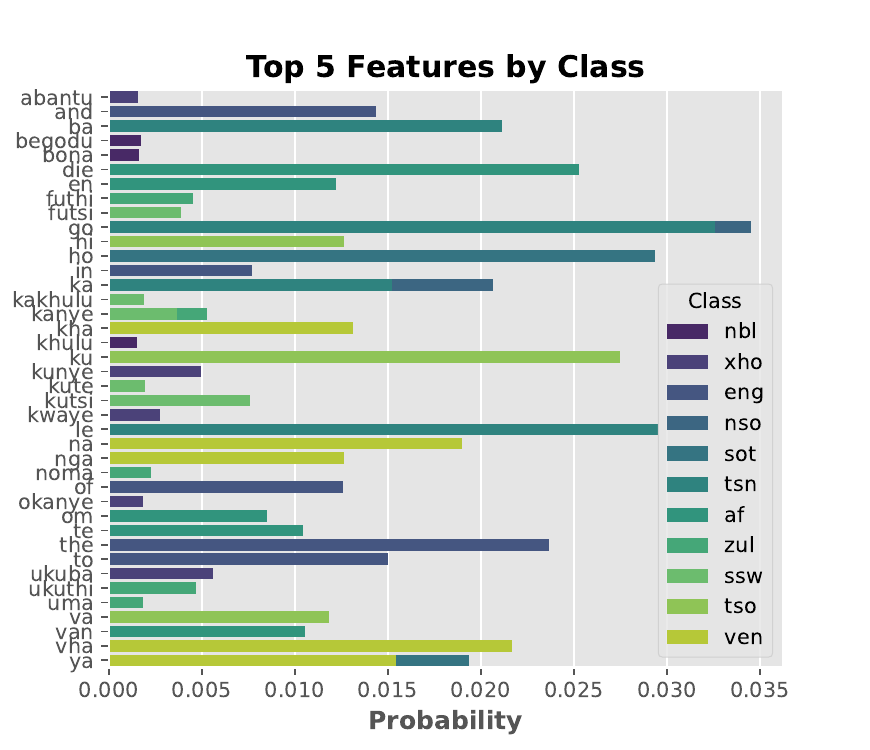}}{
        \caption{Top 5 important features per class from Naive Bayes}
        \label{important_features}
    }
\end{figure}

\begin{figure}[!ht]
\begin{center}
\includegraphics[scale=0.5]{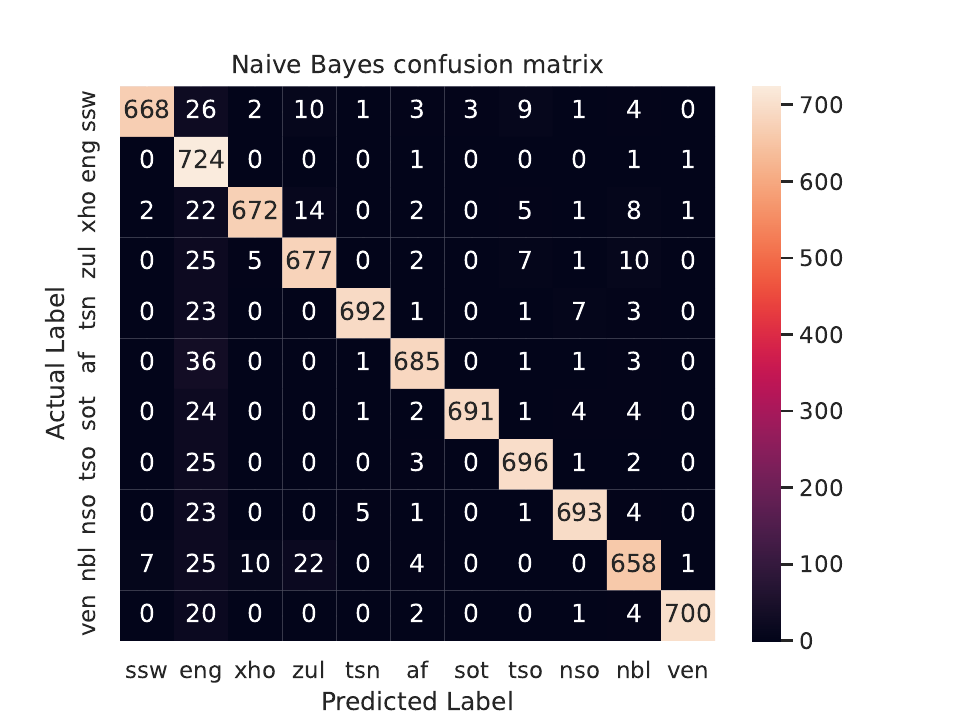} 
\caption{Accuracy of Naive Bayes Classifier.}
\label{naivebayesconf1}
\end{center}
\end{figure}

Naive Bayes have been the default standard for various LID tasks such as code-switching detection, dialect discrimination, word-level language detection, and e.t.c. \cite{dube2019language, jauhiainen2019discriminating}. In this study, we experimented with the multinomial Naive Bayes Classifier (NBC) implementation from Python's scikit-learn. With NBC, we were able to extract discriminating features per language, supporting model prediction (Figure \ref{important_features}), and significantly improved on N-gram models (see confusion matrix in Figure \ref{naivebayesconf1}). This highlighted important feature correlation, especially for related languages, which explains why it is challenging to discriminate among closely related languages. Moreover, this highlights the importance of lexicon-driven approaches for language filtering mentioned in \citet{caswell2020language} as alternative measures to mitigate these ambiguities.  

% % --------------------Bi-Figure---------------------

% --------------------NBC-Figure---------------------

\textbf{Naive Bayes Classifier experimental setup} We experimented with a TF-IDF vectorizer to generate input features. For this, we used the character bi-gram, tri-gram, quad-gram, and the 3 types combined as consecutive subwords to generate TF-IDF features. We also generated word level input features using CountVectorizer. We used a multinomial version of the Naive Bayes classifier with mostly default parameters from scikit-learn (except the alpha parameter where we tested $\alpha = 0.0001, 1.0 $, where $\alpha = 1.0$ performed better). Finally, we trained Support Vector Machine (SVM), K Nearest Neighbor (KNN), and Logistic Regression with the same input features and their scikit-learn default parameters to compare performance outcomes with NBC.

\subsubsection{Pre-trained Multilingual Models}
This study explored a diverse set of massively pre-trained multilingual models: mBERT, XLM-r, RemBERT, and their Afri-centric counterparts: AfriBERTa, Afro-XLMr, AfroLM, and Serengeti due to their enhanced text processing capabilities and their ability to handle low-resourced languages with complex linguistic nuances \cite{devlin2018bert, conneau2019unsupervised,ogueji-etal-2021,alabi-etal-2022, dossou-etal-2022-afrolm, adebara-etal-2023-serengeti}.
% Table \ref{}, shows a model versus a number of African languages used in pertaining, together with the number of South African languages included.

\textbf{Large pre-trained multilingual models experimental setup}
Following setups in \cite{adelani2023masakhanews, dione2023masakhapos}, we used a batch size of 16, a learning rate of $2\myexp^{-5}$, 20 epochs, save step of 10000, and sequences cut-off of 200 for all models. We ran our experiments five times with different seeds \{ 1,., 5\} and reported the average results.
% \subsection{Experimental Design}

\section{Results}
\subsection{Baselines}
\begin{figure*}[h]
\floatsetup{captionskip=12pt} % Adjust the value as needed
\centering
    \begin{subfloatrow}[3]
        \centering 
        \ffigbox[\hsize]{\caption{Unigram}  \label{ablation-ml-models1}}{\includegraphics[width=0.3\textwidth]{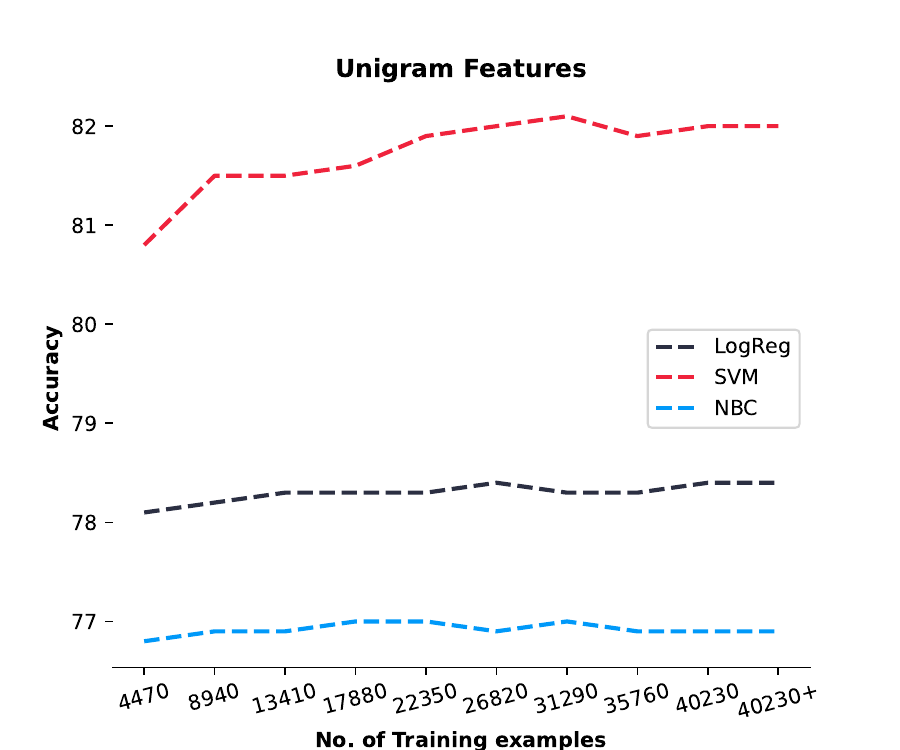}} 
        \ffigbox[\hsize]{\caption{Bi-gram} \label{ablation-ml-models2}} {\includegraphics[width=0.3\textwidth]{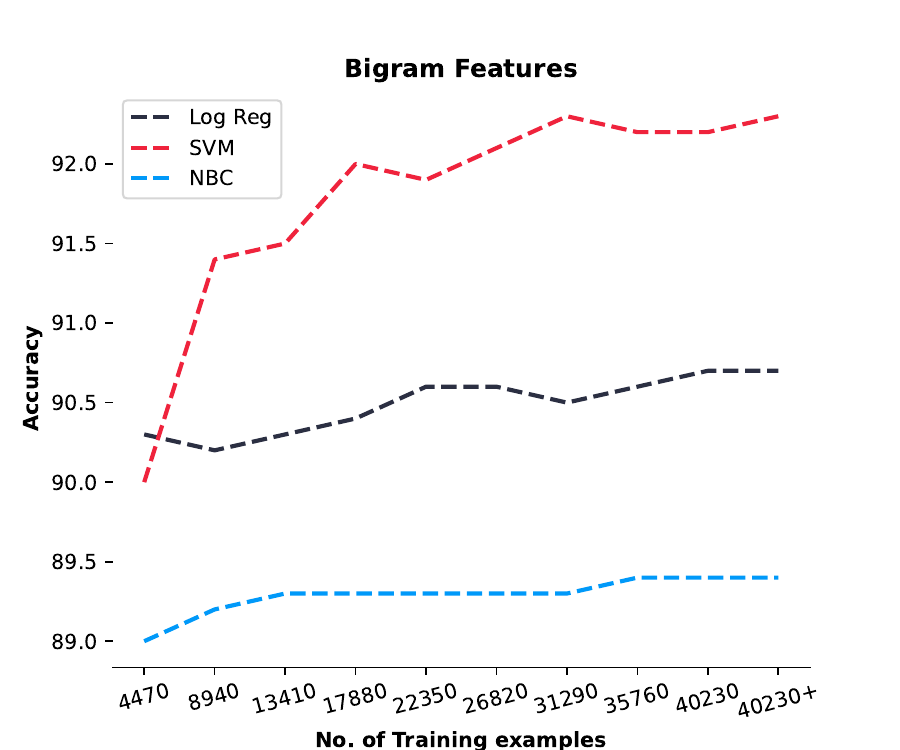}} 
        \ffigbox[\hsize]{\caption{Tri-gram}\label{ablation-ml-models3}}{\includegraphics[width=0.3\textwidth]{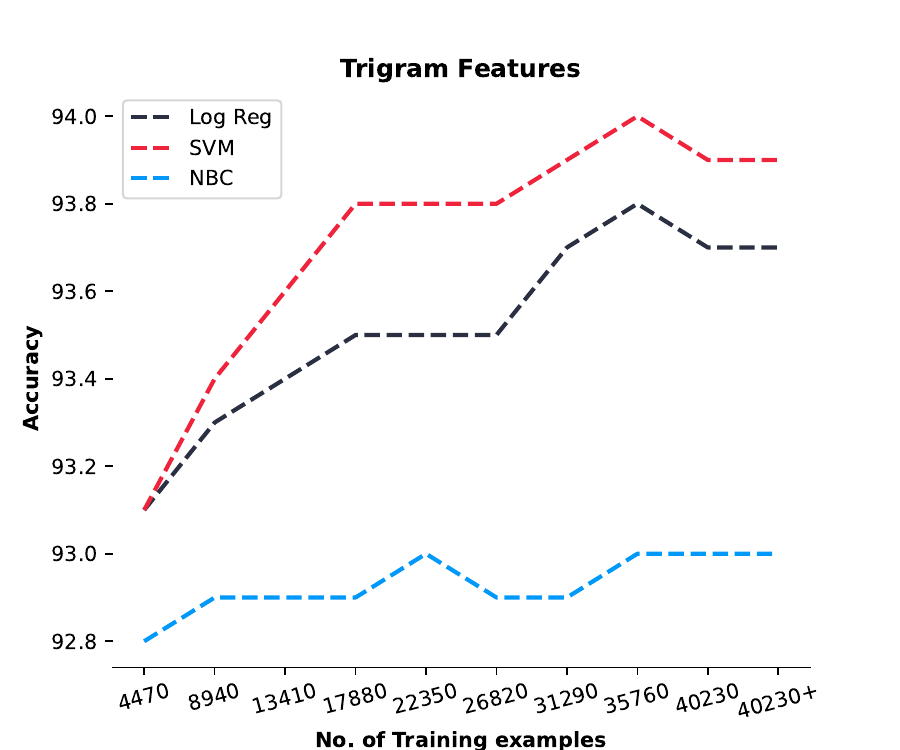}}
    \end{subfloatrow}
    \begin{subfloatrow}[3]
        \centering 
        \ffigbox[\hsize]{\caption{Quad-gram}  \label{ablation-ml-models4}}{\includegraphics[width=0.3\textwidth]{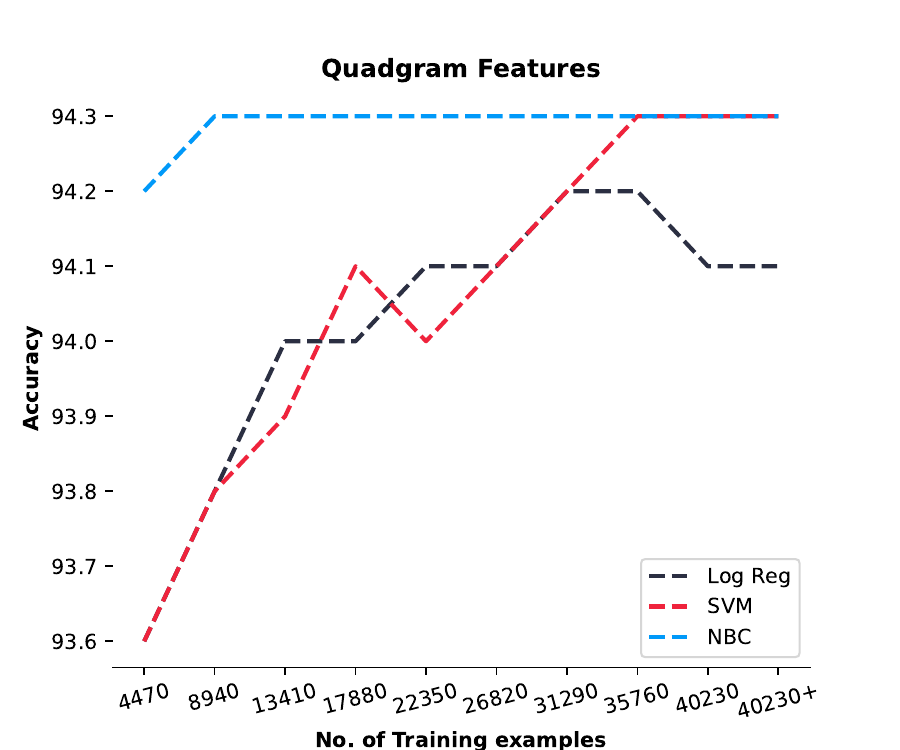}} 
        \ffigbox[\hsize]{\caption{\mbox{N-grams} Comb} \label{ablation-ml-models5}} {\includegraphics[width=0.3\textwidth]{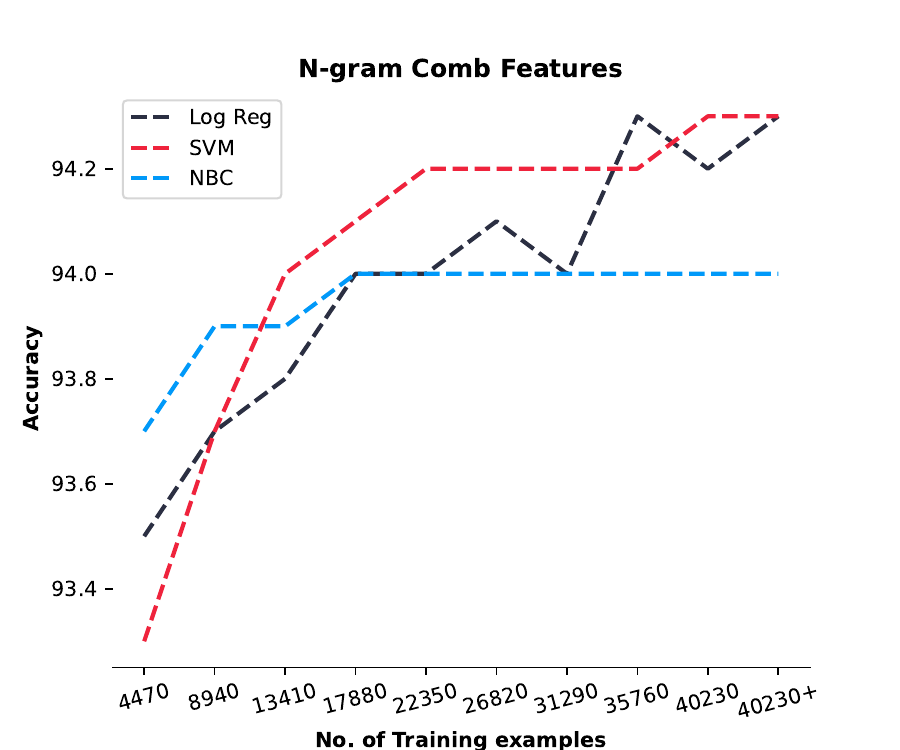}} 
        \ffigbox[\hsize]{\caption{Word-level}\label{ablation-ml-models6}}{\includegraphics[width=0.3\textwidth]{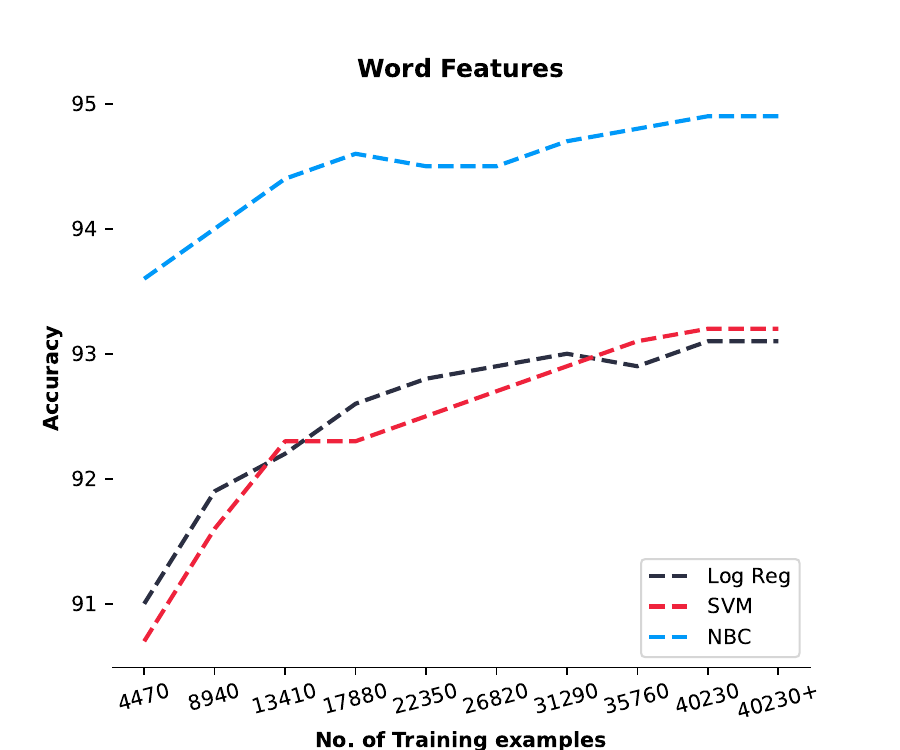}}
    \end{subfloatrow}
    \caption{Comparison of cross-lingual transfer models.}
\label{ablation-ml-models}
\end{figure*}

Table \ref{Baseline Results}, shows results for baseline models Bi-gram, Tri-gram, Quad-gram, N-gram combined (N-gram Comb) -- which uses bi-, tri-, and quad- -grams combined, and Naive Bayes Classifier (NBC) with the same character \mbox{N-grams}. Naive Bayes with word-level features outperform the rest of the baseline models. Interestingly, for NBC, increasing the character spans improves the performance of the classifier. Figure \ref{ablation-ml-models1}, \ref{ablation-ml-models2}, \ref{ablation-ml-models3}, \ref{ablation-ml-models4}, \ref{ablation-ml-models5}, and \ref{ablation-ml-models6} depicts the impact of increasing the data size on models NBC, Support Vector Machine (SVM),  and Logistic Regression (Log Reg) on various training features -- uni-grams, bi-grams, tri-grams, quad-grams, \mbox{N-grams} combined, and word-level features derived using TF-IDF respectively. NBC, SVM, and Log Reg show improved performance with the change in input features while the training size shows gradual improvement in accuracy. KNN was also tried, however, the model showed abysmal performance across all features except for Bi-gram input features and was therefore omitted from the plots.  

% Figure with sublots of feature difference

\begin{figure}[!ht]
    \centering
    \ffigbox[\FBwidth]{\includegraphics[scale=0.4]{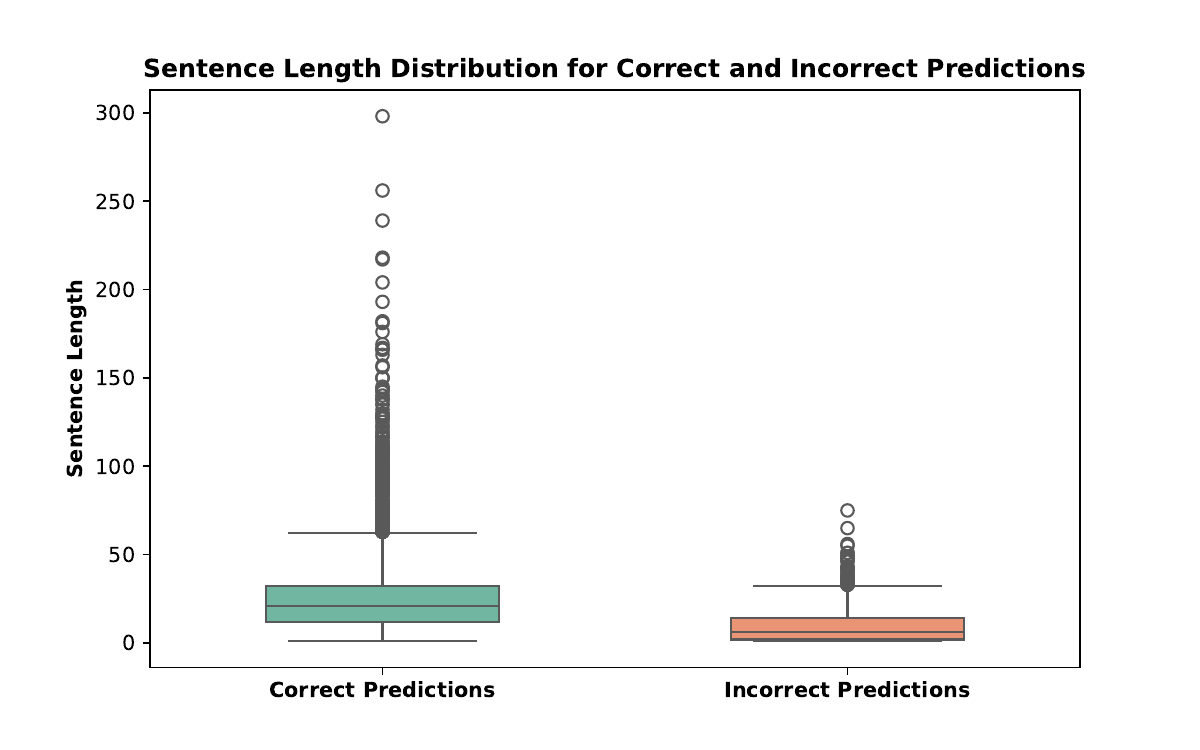}}{
        \caption{Box diagram depicting sentence length of correctly predicted and incorrectly predicted sentences.}
        \label{boxplotsentlengvsprediction.1}
    }
\end{figure}

\begin{figure}[!ht]
\begin{center}
\includegraphics[scale=0.4]{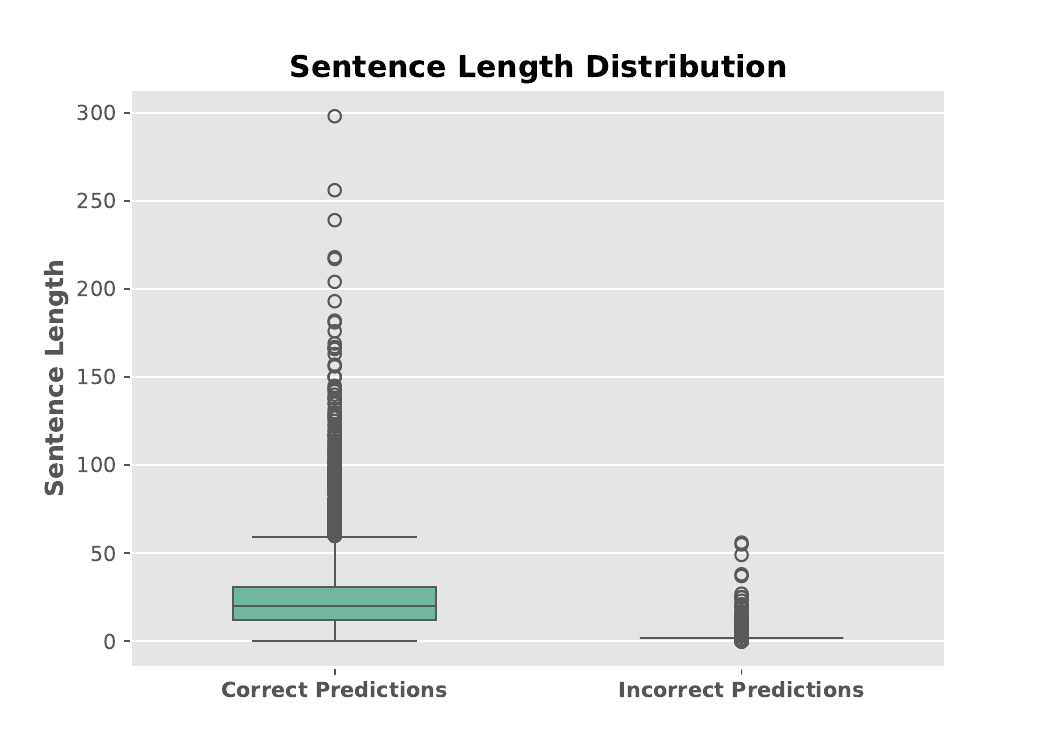} 
\caption{Incorrectly and correctly NBC classified sentence lengths}
\label{nbc_violin}
\end{center}
\end{figure}

\begin{figure}[!ht]
\begin{center}
\includegraphics[scale=0.4]{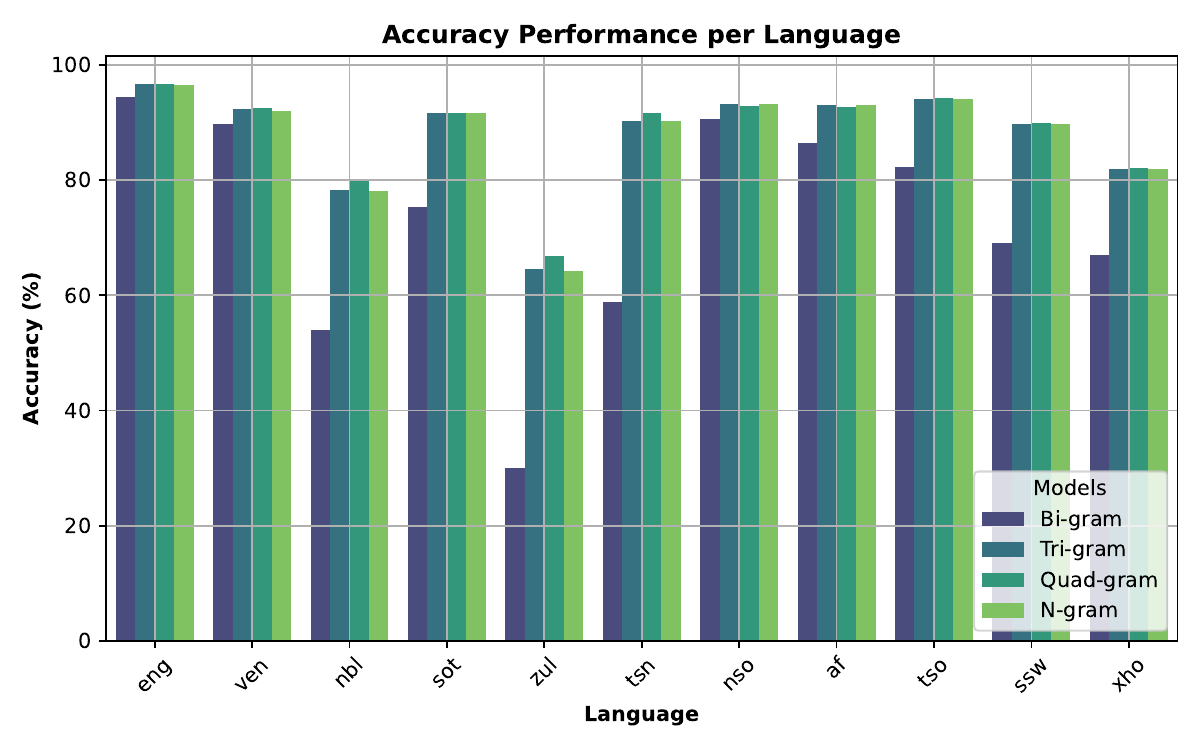} 
\caption{Accuracy score per language using \mbox{N-grams}}
\label{perlanguagescores}
\end{center}
\end{figure}

\begin{figure}[!ht]
\begin{center}
\includegraphics[scale=0.5]{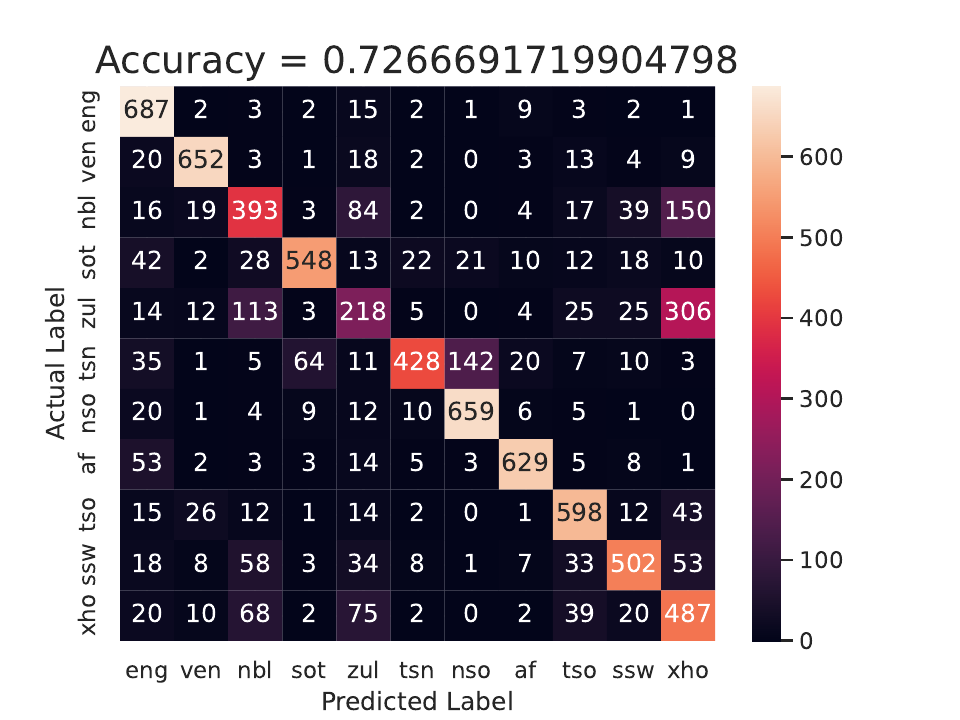} 
\caption{Bi-gram Confusion matrix on Vuk test data}
\label{conffig.1}
\end{center}
\end{figure}

\begin{figure}[!ht]
\begin{center}
\includegraphics[scale=0.5]{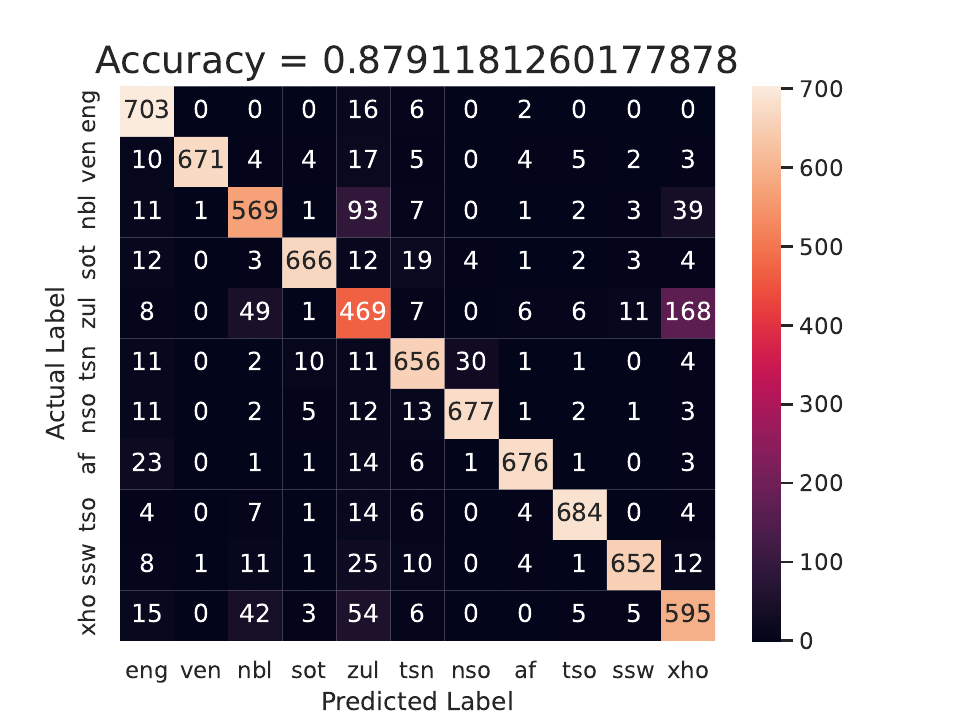} 
\caption{Tri-gram Confusion matrix on Vuk test data}
\label{conffig.2}
\end{center}
\end{figure}

\begin{figure}[!ht]
\begin{center}
\includegraphics[scale=0.5]{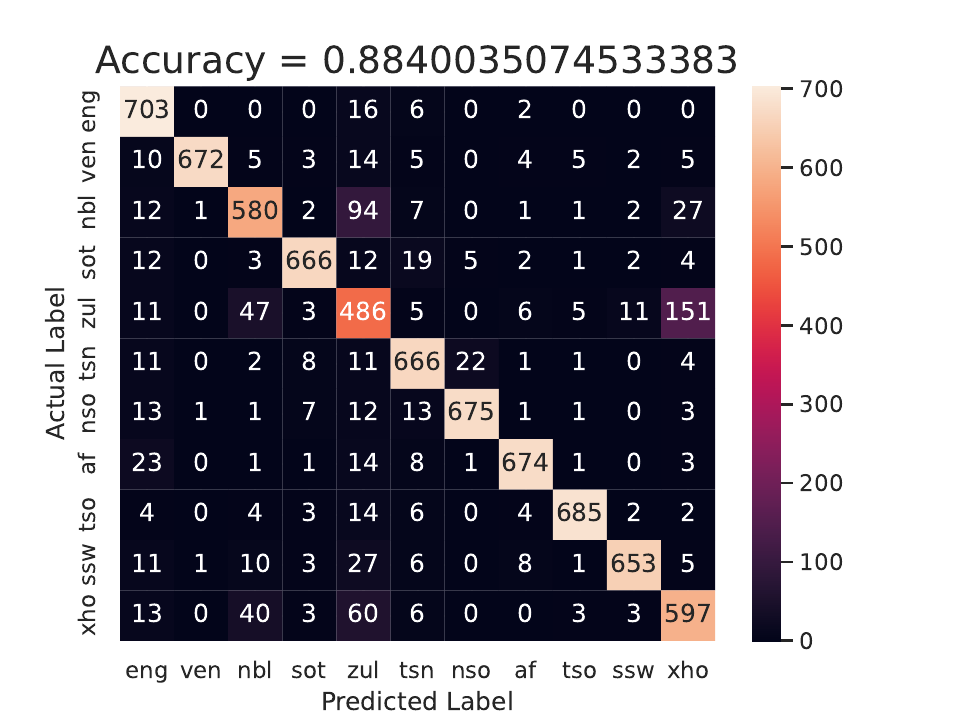} 
\caption{Quad-gram Confusion matrix on Vuk test data}
\label{conffig.3}
\end{center}
\end{figure}

\begin{figure}[!ht]
\begin{center}
\includegraphics[scale=0.5]{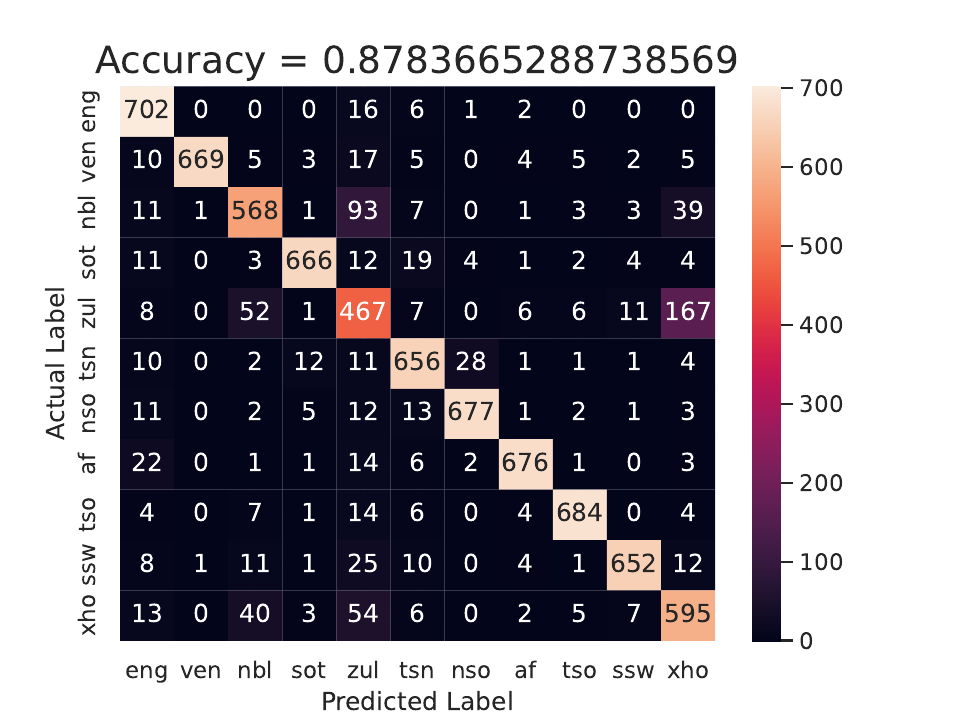} 
\caption{Accuracy of N-gram type (Bi-gram, Tri-gram, Quad-gram) combined}
\label{conffig.4}
\end{center}
\end{figure}

In the N-gram class, the Quad-gram ranking outperforms the rest of the N-gram-based models. Figure \ref{boxplotsentlengvsprediction.1}, depicts the impact of sentence length on N-gram models performance. This shows that the group of N-gram models struggles to classify shorter sentences, while NBC performs slightly better with them (Figure~\ref{nbc_violin}). This may be due to shorter sentences not carrying enough signal information for \mbox{N-grams} to discriminate across all languages as mentioned in \citet{haas2020discriminating}. Additionally, N-gram-based models depict inconsistent performance across languages, where improved performance is achieved for select languages and for a specific N-gram type (E.g Bigram -- eng, ven, af, e.t.c, Tri-gram -- eng, tso, nso, e.t.c), while other languages underperform (e.g zul, isiNdebele (nbl)) (see Figure \ref{perlanguagescores}). Furthermore, the complexity of LID is exacerbated by closely related languages (see confusion matrix in \crefrange{conffig.1}{conffig.4}). While varying dataset size, and character N-gram choices slightly improve performance on distinguishing among closely related languages (Figure \ref{ablationonvuk}), it does not add any significant improvement on a per-language basis (see Figure \ref{perlanguagescores}), where languages such as isuZulu (zul) are showing no further improvement. For this, we explore large pre-trained multilingual models for automatic LID in the next subsection. 

% -------------------Sentence length -Figure---------------------
% \begin{figure}[!ht]
% \begin{center}
% \includegraphics[scale=0.4]{images_and_plots/_box.pdf} 
% \caption{}
% \label{boxplotsentlengvsprediction.1}
% \end{center}
% \end{figure}

% --------------------Sentlen Violin plot-Figure---------------------

% -------------------Per language accuracy -Figure---------------------

% --------------------Bi-Figure---------------------

% --------------------Tri-Figure---------------------

% --------------------Quad-Figure---------------------

% --------------------N-gram-Figure---------------------

% --------------------Table---------------------
\begin{table}
\centering
\begin{tabular}{lllll}
\hline
\textbf{Baseline} & Acc & Prec & Rec  & F1 \\
\hline
Vuk & & & &  \\
Bi-gram &72.7 & 73.5& 72.6 & 72.3 \\
Tri-gram & 87.9 &88.4 & 87.9 & 88.1 \\
Quad-gram &88.4 & 88.9& 88.4 & \textbf{88.5} \\
N-gram (Comb) & 87.8&88.3 & 87.8 & 88.0 \\
NBC (word-level) & 94.5&  \textbf{95.2} &94.5 & 94.6   \\
NBC (2) & 90.2&90.7 & 90.2& 90.4  \\
NBC (3) & 93.4 &93.8 &93.4& 93.5   \\
NBC (4) &\textbf{94.4}  &94.8 & 94.4& 94.5   \\
NBC (Comb) &94.0 & 94.5& \textbf{94.0}& 94.1  \\
K NN (2) & 85.0 &85.0 & 85.0& 85.0 \\
Log Reg (4)& 94.0 & 95.0& 94.0 & 94.0\\
SVM (4 \& 2-4) &94.0 & 95.0& 94.0 & 94.0 \\
% \hline
% NCHLT & & & &  \\
% Bi-gram & & & &  \\
% Tri-gram & & & &  \\
% Quad-gram & & & &   \\
% N-gram (Comb) & & & &   \\
% NBC (2) & & & &   \\
% NBC (3) & & & &   \\
% NBC (4) & & & &   \\
% NBC (2-4) & & & &   \\
% \hline
% Vuk + NCHLT & & & &  \\
% Bi-gram & & & &  \\
% Tri-gram & & & &  \\
% Quad-gram & & & &   \\
% N-gram (Comb) & & & &   \\
% NBC (2) & & & &   \\
% NBC (3) & & & &   \\
% NBC (4) & & & &   \\
% NBC (2-4) & & & &   \\
\hline
\end{tabular}
\caption{\label{Baseline Results} Baseline performance evaluation using Accuracy (Acc), F1 score (F1), Precision (Prec), and Recall (Rec). K Nearest Neighbor (K NN), Logistic Regression (LR), and Support Vector Machine (SVM) are reported with best feature inputs bi-gram (2), quad-grams (4), and combinations (2-4) respectively.}
\end{table}

\subsection{Pre-trained Multilingual Models}
Table \ref{pre-trained models Results} reports the accuracy (Acc), precision (Prec), recall (Rec), and F1 score (F1) of pre-trained multilingual models: mBERT, XLM-r, RemBERT; Afri-centric pre-trained models: AfriBERTa, Afro-XLMr, AfroLM, and Serengeti; publicly available LID tools covering South African languages: Compact Language Detector (CLD) version 3 (V3), AfroLID \cite{adebara2022afrolid}, GlotLID \cite{kargaran2023glotlid}, and OpenLID \cite{burchell2023open}; and our proposed lightweight BERT-based architectures: za-BERT-lid, and DistilBERT.

Pre-trained-multilingual models show impressive results for this task, with over 90\% average accuracy. Serengeti outperforms the rest of the models with an average accuracy of 98 \%, while mBERT is the least-performing model with an average accuracy of 96 \% ($\approx$ 2 points difference). Most importantly, the group of Afri-centric models out-performs the largely pre-trained multilingual models with the best model (XLMr-large) in this category performing slightly worse than the lowest performing model (AfroLM) in the Afri-centric group. Moreover, our proposed za-BERT-lid, and DistilBERT perform on par with the best-performing model ($\approx$ 2 points difference) despite them being much smaller in size.   

On the other hand, available LID tools show impressive and incremental results. For these models, GlotLID outperforms the rest of the sampled models in this study. This may be due to GlotLID being trained on Vuk data, giving the model an unfair advantage over others. Despite this, analyses of the predictions show that the compared models are not completely wrong, as they often struggle with closely related languages such as Sotho-Tswana language family \{nso, sot, tsn\}, and Nguni languages \{xho, zul, ssw, and nbl\}. Perhaps to remedy this, the training of LID models should prioritize precision as a metric of evaluation. Noticeably, but not alarming, the LID tools also predict unrelated languages from their training list, which perhaps highlights the need for a more focused approach rather than including many languages at once. However, we feel this claim needs further justification and we will consider this in future work.

\begin{table}
\centering
\begin{tabular}{lllll}
\hline
\textbf{Model} & Acc & Prec & Rec & F1  \\
\hline
PLM & & & &  \\
mBERT &96.7 & 96.7& 96.6 & 96.7\\
XLMr-base &97.1 & 97.1 & 97.1& 97.1\\
XLMr-large & 97.3& \textbf{97.3} & 97.3& 97.3\\
RemBERT &97.1&97.1 & 97.1& 97.1 \\
\hline
Afri-centric & & &  & \\
AfriBERTa & 97.6 & 97.6 &  97.6& 97.6 \\
Afro-XLMr-base &97.7 & 97.8 & 97.7&97.7\\
Afro-XLMr-large &98.0 & \textbf{98.0} & 98.0&98.0\\
AfroLM &97.4& 97.5 & 97.4 & 97.4\\
Serengeti &98.3 & \textbf{98.3} & 98.3&98.3 \\
\hline
LID Tools & & & &  \\
CLD V3 & 40.2 & 33.6 & 40.2& 35.7\\
AfroLID & 66.1 &72.1 & 66.1& 64.2 \\
OpenLID & 80.8 & 71.7 & 80.8& 75.0  \\
GlotLID & 97.5  & \textbf{98.3} & 97.5& 97.9 \\
\hline
Lightweight & & &  & \\
za-BERT-lid & 96.8 & 96.8 & 96.8& 96.8\\
DistilBERT & 96.2& 96.2 & 96.2 & 96.2\\
\hline
\end{tabular}
\caption{\label{pre-trained models Results} Performance evaluation scores of pre-trained multilingual models, available LID tools, and lightweight BERT-based models averaged over 5 runs per metric.}
\end{table}

\subsection{Cross-domain evaluation}
We also wanted to test our model on cross-domain datasets to inspect their generalization capabilities. We simulated this by training with Vuk data and tested it on NCHLT, and vice versa. Table \ref{cross-domain-validation} reports the performance of pre-trained models for examining the cross-domain evaluation theory.  This table shows that the performance of the multilingual models trained with Vuk and tested with NCHLT dropped by approximately (4\%-5\%) across all models. In contrast, training with NCHLT and testing with Vuk showed performance improvements. This could be due to NCHLT having more training examples, and a large vocabulary (see Table \ref{corpora-statistics}) allowing the model to learn more nuanced representations. Notably, larger models show better performance over smaller models for this task.

\begin{table}
\centering
\begin{tabular}{lcc} 

\hline
\textbf{Model} & Vuk Test & NCHLT Test   \\
\hline
Vuk Trained & &   \\
mBERT &- & 91.0  \\
XLMr-base & -&  91.4 \\
XLMr-large & -&  92.2 \\
RemBERT & -& 92.3  \\
AfriBERTa & -& 92.1  \\
Afro-XLMr-base & -& 93.6  \\
Afro-XLMr-large &- & \textbf{94.1}  \\
AfroLM &- & 91.8  \\
Serengeti & -&  \textbf{94.9} \\
za-BERT-lid & -& 91.3  \\
DistilBERT &- &  90.9 \\
\hline
NCHLT Trained & &   \\
XLMr-base & 95.6 & 93.2 \\
Afro-XLMr-base &96.3 & 93.6  \\
% Afro-XLMr-large & 97.0 &   \\
Serengeti & 97.7 & \textbf{94.8}   \\
% DistilBERT & &   \\ 
\hline

\end{tabular}
\caption{\label{cross-domain-validation} Cross-domain evaluation of models trained with Vuk and tested with NCHLT and vice-versa. Reported in F1 score averaged over five runs}
\end{table}

\section{Discussions}
Ensuring the development of robust LID detection systems remains a critical research area with implications on many NLP tasks. Importantly, the availability of reliable LID systems ensures accurate reporting on the state of low-resourced languages \cite{kreutzer2022quality}.

On the side of model performance, baseline techniques such as Naive Bayes, Support vector Machines, and Logistic Regression seem to be performing quite well on the task of sentence-level language identification. We recommend these models for further research for high-level LID, compared to large pre-trained multi-lingual models which require specialized computing resources such as GPUs, to accelerate training. However, we deem such trade-offs to require more research, especially in complex LID subtasks such as code-switching, or similar language discrimination.

We also, highlight the importance of evaluation metric selection as we have observed that most of the LID tools explored in this study are not completely wrong, but rather have challenges discriminating among closely related languages. Therefore, we recommend precision as an evaluation metric for LID to be further investigated.

% Misguidance to resource availbility

\section{Conclusion}
Language Identification remains a critical study area for the widespread inclusion of many low-resourced languages into the booming technology space. In this study, we experimented with statistical approaches, traditional machine learning techniques, the recent advanced pre-trained multilingual models, as well as LID tools publicly available (covering a wide range of African languages) on the task of LID for 11 South African language discrimination. We were able to shed light on the approaches showing promising results in the South African language context and made suggestions for future directions. Concretely, we showed that the Naive Bayes algorithm performs surprisingly well for LID and warrants further exploration and exploitation, especially given its cheap-compute advantage. Finally, we compared publicly available pre-trained models and showed that context-exposed models have an edge over other context-oblivious multilingual models, where context refers to the language. We released our models on \href{https://huggingface.co/spaces/dsfsi/dsfsi-language-identification-spaces}{HuggingFace} and code with datasets on \href{https://github.com/dsfsi/za-lid/}{GitHub}.

\section{Limitations}

In this study, we did not explore any use of word embeddings for language identification. Word embeddings played in crucial role in the development of language technologies, and it would have been interesting to experiment with them. However, such resources are not readily available for many low-resourced languages. 

Aside from experimenting and getting results for other traditional models such as Logistic regression, K Nearest Neighbor, and Support Vector Machines, it would have been interesting to develop and experiment with deep neural networks such as multi-layered perceptions, and convolutional neural networks. As universal approximators, these models tend to produce desirable results, with the caveat of requiring time for hyper-parameter tuning. 

This study did not extensively explore the impact k (used 50 for this study), which is the count of the \mbox{N-grams} list used to calculate the ranking. However, we aim to explore this extensively in future works.

It is known that LID techniques tend to overfit to domain data, and therefore it would have been interesting to create free-text data created by humans and test the generalization capabilities of the developed models on human-generated text.  

Recent studies have focused on resource-conscience alternatives for either compute efficiency, parameter reduction, etc. It would have been interesting if this work would have explored the recently active approaches focusing on smaller models utilizing parameter transfer, and adaptations \cite{kumar2023transformer}. However, these techniques require intense hyper-parameter selection and tuning, and slightly longer training times, which was not in the scope of this study.

Finally, we aim to incorporate BANTUBERT \footnote{https://huggingface.co/dsfsi/BantuBERTa}, and zaBANTUBERT \footnote{https://huggingface.co/dsfsi/zabantu-xlm-roberta} models trained with monolingual South African corpora in our future work.

\section*{Acknowledgements}
The authors would like to thank the ABSA chair of Data Science for funding and support. We also like to express our gratitude to Matimba Shingange and Michelle Terblanche for their input and reviews of the paper. Finally, the authors would like to thank the  Data Science for Social Impact Research lab  for their continued support.

\bibliography{anthology,custom}
\bibliographystyle{acl_natbib}
% \bibliographystyle{anthology, plain} % or any other style you prefer
% \bibliography{acl_natbib}
% \appendix

% \section{Example Appendix}
% \label{sec:appendix}

% This is a section in the appendix.

\end{document}